# Suspicious and Anomaly Detection

Monali Ahire, Devarshi Borse, Amey Chavan, Shubham Deshmukh, Favin Fernandes

Department of Electronics and Telecommunication, Vishwakarma Institute of Technology, Pune

monali.ahire18@vit.edu , devarshi.borse18@vit.edu , amey.chavan18@vit.edu , shubham.deshmukh18@vit.edu , favin.fernandes18@vit.edu

Abstract: In this project we propose a CNN architecture to detect anomaly and suspicious activities; the activities chosen for the project are running, jumping and kicking in public places and carrying gun, bat and knife in public places. With the trained model we compare it with the pre-existing models like Yolo, vgg16, vgg19. The trained Model is then implemented for real time detection and also used the. tflite format of the trained .h5 model to build an android classification.

## I. Introduction

Detection of suspicious human actions in automated video surveillance applications, is of great practical importance. Those kind of unusual activities in human is very difficult to acquire and classify to predict. In our proposed work, automatic tracking and detecting unusual movement's problems in closed circuit videos was resolved.

Students, faculty, staff and visitors can help protect our community by learning to recognize and report suspicious activity. Prompt and detailed reporting can help prevent crimes or terrorist attacks.

Suspicious behavior or activity can be any action that is out of place and does not fit into the usual day-to-day activity of our campus community. For example, you see someone looking into multiple vehicles or homes or testing to see if they are unlocked. Or perhaps you are worried about how your roommate has been acting differently or concerned about the behavior of a co-worker.

These are always suspicious activities:

- Person screaming, cries for HELP or POLICE.
- Loud or obscene shouting indicating a disturbance.
- An explosion or gunshot.
- The sound of breaking glass.
- Someone trying to break into a building.
- Someone tampering with a motor vehicle.
- Person(s) publicly displaying weapons.
- Smashed doors or windows.

During the course of your job you may have to challenge a suspicious person. "Challenge" means to tactfully and politely greet the individual and offer your assistance. If a person is here for legitimate reasons, he or she will appreciate the attention that you show them. Someone with the intent of engaging in criminal activity does not want attention drawn to them. It is not your role to replace law enforcement or security.

Never enter into a situation where you feel unsafe. If you feel uncomfortable challenging a suspicious person, or if your suspicions continue after making contact, report the situation to law enforcement or security immediately.

Here are common guidelines to follow when approaching a suspicious person, regardless of the circumstances:

- Make eye contact and politely greet the person: "Good evening, how can I help you?"
- Act with discretion and use tact.
- Politely inquire whether the person is a visitor, student, or employee.
- Do not accuse them or speculate as to what they might be doing.

- Do not threaten or intimidate.
- If necessary, contact security or your supervisor.

To achieve anomaly detection, one of the most widespread method is using the videos of normal events as training data to learn a model and then detecting the suspicious events which would do not fit in the learned model. For example, human pose guesstimate is used in applications including video surveillance, animal tracing and actions understanding, sign language recognition, advanced human-computer interaction, as well as marker less motion capturing. Low cost depth sensors consist of limitations like limited to indoor use, and their low resolution and noisy depth information make it difficult to estimate human poses from depth images. Hence, we are to using neural networks to overcome these problems. Anomalous human activity recognition from surveillance video is an active exploration part of image processing and computer visualization

For this project we identify 6 classes for the initial detection of activities and then propose another architecture just to classify the objects rather than the activities.

## II. Literature Review

In this project [1] the hierarchical approach is used to detect the different suspicious activities such as loitering, fainting, unauthorized entry etc. This approach is based on the motion features between the different objects. First of all, the different suspicious activities are defined using semantic approach. Then the object detection is done using background subtraction. The detected objects are then classified as living (human) or nonliving (bag). These objects are required to be tracked which is done using correlation technique. Finally using the motion features & temporal information the events are classified as normal or suspicious. As the semantic based approach is used computational complexity is less and the efficiency of the approach is more.

In this project [2], different systems are in place which helps to differentiate various suspicious behaviors from the live tracking of footages. The most unpredictable one is human behavior and it is very difficult to find whether it is suspicious or normal. Deep learning approach is used to detect suspicious or normal activity in an academic environment, and which sends an alert message to the corresponding authority, in case of predicting a suspicious activity. Monitoring is often performed through consecutive frames which are extracted from the video. The entire framework is divided into two parts. In the first part, the features are computed from video frames and in second part, based on the obtained features classifier predict the class as suspicious or normal.

In this project [3], a surveillance system that employs Human Activity Recognition techniques which can efficiently decide if the objective individual is an ordinary individual or a suspicious individual can be deployed. It is likewise expected that establishing detection systems can act as a hindrance against crime. This paper proposes a surveillance system that utilizes YOLO and ResNet for detecting suspicious individuals and activities.

This paper [4] proposes an intelligent system, which utilizes an expert video-surveillance tool for monitoring of potentially suspicious and criminal activities in shopping malls. The system uses methods such as object detection, movement tracking and activity monitoring to robustly and efficiently track the location of objects and the subjects to identify questionable actions and activities. Along with this, an interface is developed to notify the concerned authority in real time. Thus, the system works as an assistant to the security personnel. The system uses various custom objects and person detection models to identify and establish a relationship between them.

The proposed method in [5] finds difference between two consecutive video frames in order to find the motion history image by eliminating the use of labels. We detect suspicious behavior in videos and live surveillance using a spatiotemporal architecture. The architecture for prediction of video frames is based on convolutional neural networks. Classification of normal and abnormal behavior is attained for most of the scenarios. An anomalous batch of frames within the videos is identified and an alarm is generated to alert people about the malicious activities. In this way, they can be prevented on the spot instead of taking action later.

## III. Methodology

The Dataset collected for this projects were an accumulation of open source datasets available on Kaggle. A total of 6 classes were formed.

Bat:

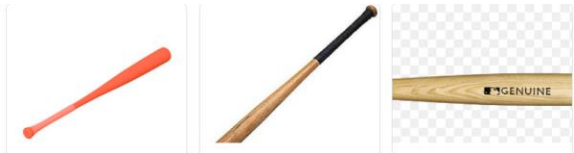

Knife:

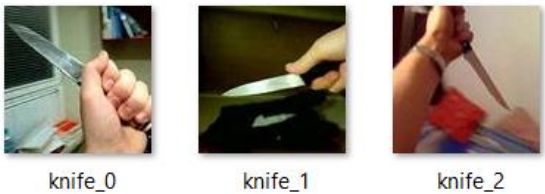

Gun:

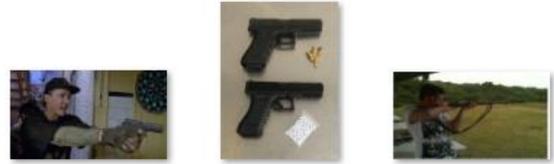

Jumping, Running & Kicking:

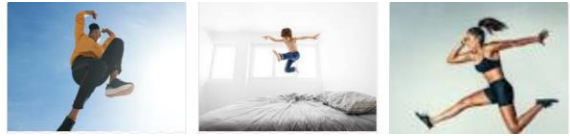

The methodoly was this project was split into 4 parts:

A) Detection of 6 classes using proposed CNN Architecture.

B) Detection of Guns using Yolo

C) Detection of Activities Gun, Bat and knife using the proposed CNN model.

D) Detection of Activities based on haarcascade Detection and CNN for Classification.

CNN

A Convolutional Neural Network or CNN is a special type of neural networks that specializes in processing data that has a grid-like topology, such as an image. Neural Network is well known for finding out patterns in various kind of datasets, convolutional is particularly used for images as image being huge chunk of data convolution operation can extract meaning information from it without losing much information and perform task quickly. Any CNN model is comprised of three basic layers: convolution layer, pooling layer, fully connected layer. In convolutional layer it performs a dot product between two matrices, where one matrix is the set of learnable parameters otherwise known as a kernel, and the other matrix is the restricted portion of the receptive field. The kernel is used for feature extraction for finding out various features in an image such as vertical, horizontal edges etc., these set of kernels are convolved with the input image. The kernel is spatially smaller than an image but is more in-depth. This means that, if the image is composed of three (RGB) channels, the kernel height and width will be spatially small, but the depth extends up to all three channels. Pooiling layers are used to extract most useful information from the convolved outputs, various other commands are used to prevent overfitting. Fully connected layers are used perform normal neural network operations to detect patterns from the outputs of previous layers. Along with convolution layers' various activation functions like relu, sigmoid, and softmax are used to formulate conv layers output.

The proposed CNN architecture used for Suspicious and Anomaly detection uses 4 Conv2d ,4 MaxPool2d, 4 Dropout, 1 GlobalAveragePooling2d, and 3 Flatten Dense layers finally giving us 36 multi class classification output using softmax. Using Adam optimizer as an optimization technique along with categorical crossentropy as an accuracy metrics the model is fitted and achieving training accuracy as 95% and testing accuracy as 94% for dangerous objects, for Activities its 91% training and 92% testing. Now the weights and biases trained through the proposed architecture are saved in keras model format (.h5 model) and later can again be converted to tflite model from which it has been used to detect sign language through mobile android application in real time.

HaarCascade

Haar Cascade classifier is an effective object detection approach which was proposed by Paul Viola and Michael Jones in their paper, "Rapid Object Detection using a Boosted Cascade of Simple Features" in 2001.

 Initially, the algorithm needs a lot of positive images (images of faces) and negative images (images without faces) to train the classifier. Then we need to extract features from it. For this, Haar features shown in the below image are used. They are just like our convolutional kernel. Each feature is a single value obtained by subtracting sum of pixels under the white rectangle from sum of pixels under the black rectangle.

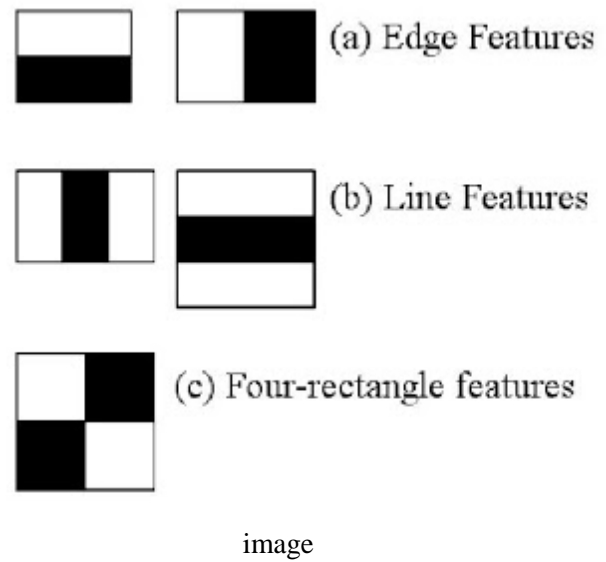

image

Now, all possible sizes and locations of each kernel are used to calculate lots of features. (Just imagine how much computation it needs? Even a 24x24 window results over 160000 features). For each feature calculation, we need to find the sum of the pixels under white and black rectangles. To solve this, they introduced the integral image. However large your image, it reduces the calculations for a given pixel to an operation involving just four pixels. Nice, isn't it? It makes things super-fast.

But among all these features we calculated, most of them are irrelevant. For example, consider the image below. The top row shows two good features. The first feature selected seems to focus on the property that the region of the eyes is often darker than the region of the nose and cheeks. The second feature selected relies on the property that the eyes are darker than the bridge of the nose. But the same windows applied to cheeks or any other place is irrelevant. So how do we select the best features out of 160000+ features? It is achieved by **Adaboost**.

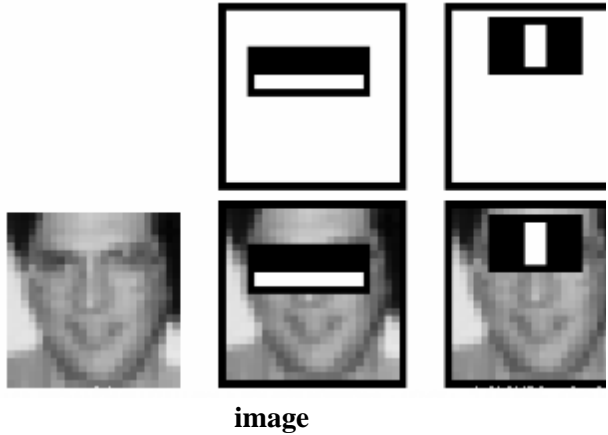

**image**

For this, we apply each and every feature on all the training images. For each feature, it finds the best threshold which will classify the faces to positive and negative. Obviously, there will be errors or misclassifications. We select the features with minimum error rate, which means they are the features that most accurately classify the face and non-face images. (The process is not as simple as this. Each image is given an equal weight in the beginning. After each classification, weights of misclassified images are increased. Then the same process is done. New error rates are calculated. Also new weights. The process is continued until the required accuracy or error rate is achieved or the required number of features are found).

The final classifier is a weighted sum of these weak classifiers. It is called weak because it alone can't classify the image, but together with others forms a strong classifier. The paper says even 200 features provide detection with 95% accuracy. Their final setup had around 6000 features. (Imagine a reduction from 160000+ features to 6000 features. That is a big gain).

So now you take an image. Take each 24x24 window. Apply 6000 features to it. Check if it is face or not.

In an image, most of the image is non-face region. So it is a better idea to have a simple method to check if a window is not a face region. If it is not, discard it in a single shot, and don't process it again. Instead, focus on regions where there can be a face. This way, we spend more time checking possible face regions.

For this they introduced the concept of **Cascade of Classifiers**. Instead of applying all 6000 features on a window, the features are grouped into different stages of classifiers and applied one-by-one. (Normally the first few stages will contain very many fewer features). If a window fails the first stage, discard it. We don't consider the remaining features on it. If it passes, apply the second stage of features and continue the process. The window which passes all stages is a face region. How is that plan!

The authors' detector had 6000+ features with 38 stages with 1, 10, 25, 25 and 50 features in the first five stages. (The two features in the above image are actually obtained as the best two features from Adaboost). According to the authors, on average 10 features out of 6000+ are evaluated per sub-window.

After successful completion of all the 4 methods the trained CNN model is saved in .h5 format which is then converted to .tflite model to build an Android Application.

## IV. Results

A) Detection of 6 classes using proposed CNN Architecture:

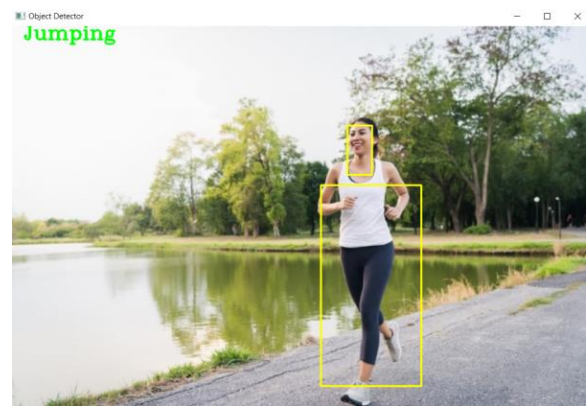

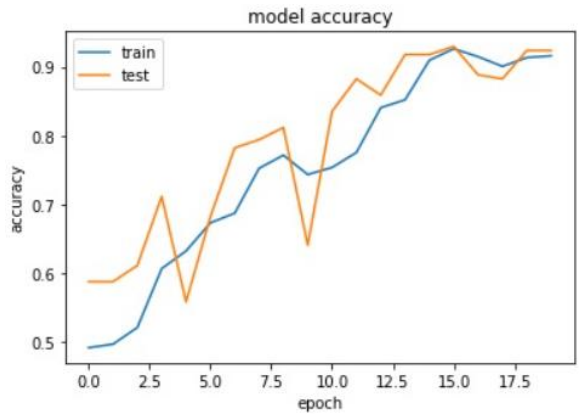

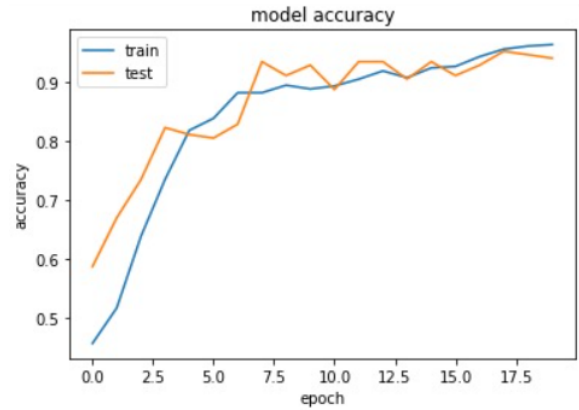

B) Detection of Guns using Yolo:

D) Detection of Activities based on haarcascade Detection and CNN for Classification:

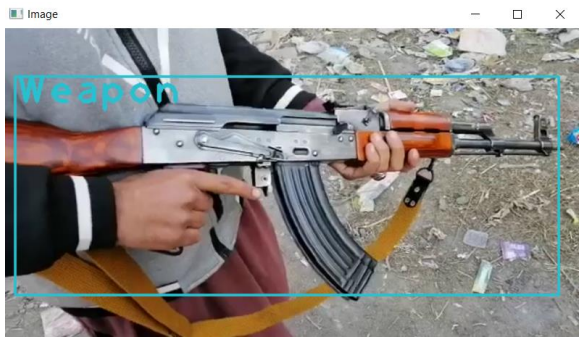

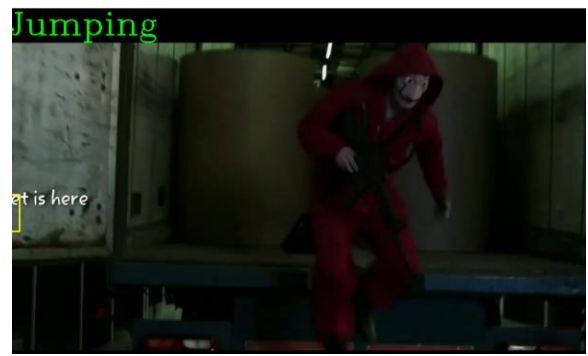

C) Detection of Activities Gun, Bat and knife using the proposed CNN model:

E) Android results

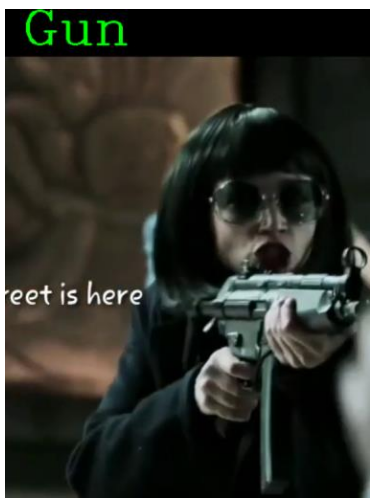

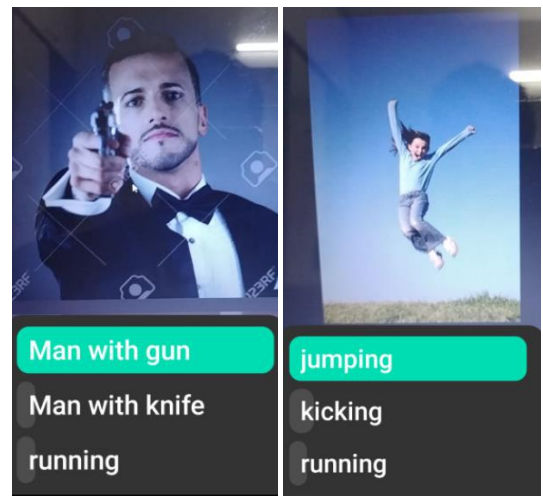

## V. Conclusion

In this project. We have gone through multiple methods in order to classify unusual activities in public places. Using HaarCascade, it is a good method for body detection but the drawback of using it is that it is computationally expensive and the rate of frames of video decreases while processing the images, due to this the CNN models cannot classify accurately the class of the image given to it. The proposed CNN model to detect both the dangerous objects and activities had an accuracy of around 92% which is fairly good for classification and recognition. We also compared the results with pre trained model architectures such as vgg16, vgg19, ResNet50, and Yolov3.